# SPATIO-TEMPORAL CO-OCCURRENCE CHARACTERIZATIONS FOR HUMAN ACTION CLASSIFICATION

*Aznul Qalid Md Sabri[1], Jacques Boonaert[2], Erma Rahayu Mohd Faizal Abdullah[1] and Ali Mohammed Mansoor[1]*
[1]Faculty of Computer Science and Information Technology, University of Malaya, Malaysia
[2]Mines Douai, Computer Science & Automatic Control Research Unit, 764 boulevard Lahure, Douai, France

E-mail: aznulqalid@um.edu.my, jacques.boonaert@mines-douai.fr, erma@um.edu.my, ali.mansoor@um.edu.my

## ABSTRACT

*The human action classification task is a widely researched topic and is still an open problem. Many state-of-the-arts approaches involve the usage of bag-of-video-words with spatio-temporal local features to construct characterizations for human actions. In order to improve beyond this standard approach, we investigate the usage of co-occurrences between local features. We propose the usage of co-occurrences information to characterize human actions. A trade-off factor is used to define an optimal trade-off between vocabulary size and classification rate. Next, a spatio-temporal co-occurrence technique is applied to extract co-occurrence information between labeled local features. Novel characterizations for human actions are then constructed. These include a vector quantized correlogram-elements vector, a highly discriminative PCA (Principal Components Analysis) co-occurrence vector and a Haralick texture vector. Multi-channel kernel SVM (support vector machine) is utilized for classification. For evaluation, the well known KTH as well as the challenging UCF-Sports action datasets are used. We obtained state-of-the-arts classification performance. We also demonstrated that we are able to fully utilize co-occurrence information, and improve the standard bag-of-video-words approach.*

## 1.0 INTRODUCTION

The characterization for human actions can be separated into being either local-based [1][2][3] or global-based [4][5]. The former is more robust as most of local-based methods are invariant to viewpoint changes, person appearance and partial occlusions [6]. Unlike the latter, accurate localization and background subtraction are not required. For these reasons, we have chosen to utilize local based characterization.

The bag-of-video-words model is considered a standard local based characterization approach [7]. In this approach, a 'vocabulary' is constructed using the centers of sets of local descriptors. These cluster centers are typically obtained through k-means clustering. One of the main issues with this approach is that it ignores spatial and temporal relationship information between local features [8]. Furthermore, there is an issue on deciding the optimal size of the vocabulary [9]. One of the contribution of this paper is that we propose a trade-off factor to help determine the size of the vocabulary.

Savarese et al. [10] introduced spatio-temporal correlograms that describe co-occurrences of video-words within spatio-temporal neighborhoods. These spatio-temporal correlograms contain useful co-occurrence information (between video-words labels) that can help distinguish, similar types of human actions (e.g. jog and run), which is often a non-trivial task.

Similar to bag-of-video-words approach, correlograms elements are vector quantized (using k-means clustering) to produce representative correlograms elements denoted as "correlations". The characterization produced is referred to as bag-of-correlations. "Correlation" in this context represents the relationship between local descriptors. In [11] we highlighted that, the usage of a discriminative type of descriptor affects the overall co-occurrence information contained within the spatio-temporal correlogram. Thus, we had chosen to use SURF based descriptors [12] in replacement of the brightness gradient descriptors used by Savarese et al. [10]. This enhances the classification rate since the final characterizations for actions contained in the different videos are more discriminative. However, a question to be answered is whether this type of characterization (i.e. bag-of-correlations) is the most profitable approach in making use of the co-occurrence information contained within the correlogram? In this paper, we argue that, in the original bag-of-correlations [10] approach there is a high possibility that important information concerning the video-words labels will be lost during the vector quantization process. This is because vector quantization is performed directly on the correlogram elements without considering the video-words labels information.

To address this problem and to fully make use of the co-occurrence information, the contributions of this paper consist of presenting several novel characterizations that are directly extracted from the spatio-temporal correlogram. The initial idea for these characterizations was briefly presented in [13].

Firstly, we propose a novel characterization for human actions by extracting a set of Haralick texture measures [14] from the correlograms. In image classification, texture measures are extracted from a co-occurrence matrix [14]. In our work, we extract similar statistical texture measures from correlograms containing co-occurrence information of local features.

Secondly, we propose another type of characterization for human action that is obtained by reducing the dimensionality of the spatio-temporal correlograms using PCA [15]. Each of the spatio-temporal correlogram is projected onto a PCA subspace that reduces the number of its dimension, preserving important information.

In this paper, we present the complete work and idea concerning these two novel characterizations. New and improved results will be presented to demonstrate that the usage of our approach is beneficial in improving the overall human action classification performance.

Both KTH [16] and the challenging UCF-Sports [5] datasets are used to evaluate our approach. From our experiments, we are able to demonstrate that our approach succeeds in classifying different action classes, and is able to achieve state-of-the-art performance. This claim is supported by a detailed experimental results. Furthermore, the benefit of our approach is that we are able to fully make use of the spatio-temporal co-occurrence information within the correlograms. Our approach also addresses the issue faced formerly (in [10] and [11]), concerning the lost of spatio-temporal co-occurrence information during vector quantization process.

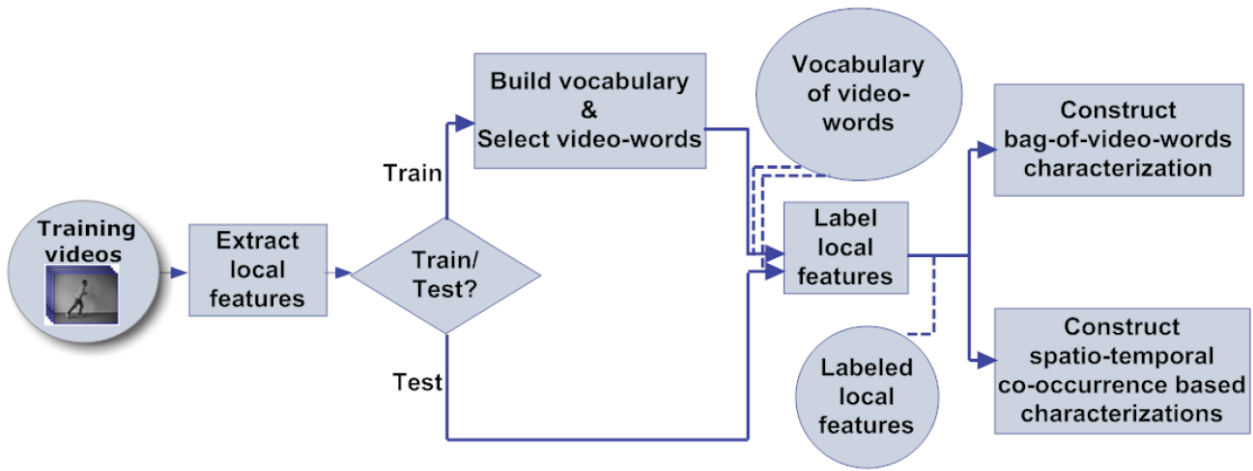

Fig. 1: Global flow of the proposed approach

Fig. 1 depicts the global flow of our proposed approach. The different processes involved in our approach are represented by the different blocks presented in Fig. 1. The entire process begins with an initial set of videos. These videos contain different actions such as walking, jogging and hand-clapping.

During training, the first blocks are the blocks dealing with extraction of local features, and the construction of vocabulary of video-words. The vocabulary will be the basis from which we construct several distinct spatio-temporal co-occurrence based action characterizations. Each characterization can be used either individually or in combinations in order to construct a SVM classifier for human actions.

The organization of this paper is based on the different blocks involved in our proposed approach.

## 2.0     CONSTRUCTION OF VOCABULARY OF VIDEO-WORDS

This section presents our implementation of the bag-of-video-words approach. Bag-of-video-words approach starts with the detection of spatio-temporal interest points (STIP) and its corresponding descriptors. STIP constitutes most salient areas in a video indicating motion, while a descriptor is a patch surrounding the STIP that stores the motion information. Next, we build a vocabulary of video-words and perform video-words selection. Finally, a histogram is computed based on the frequency of video-words labels occurring in a video sequence containing human action. This histogram is what we refer to as bag-of-video-words characterization.

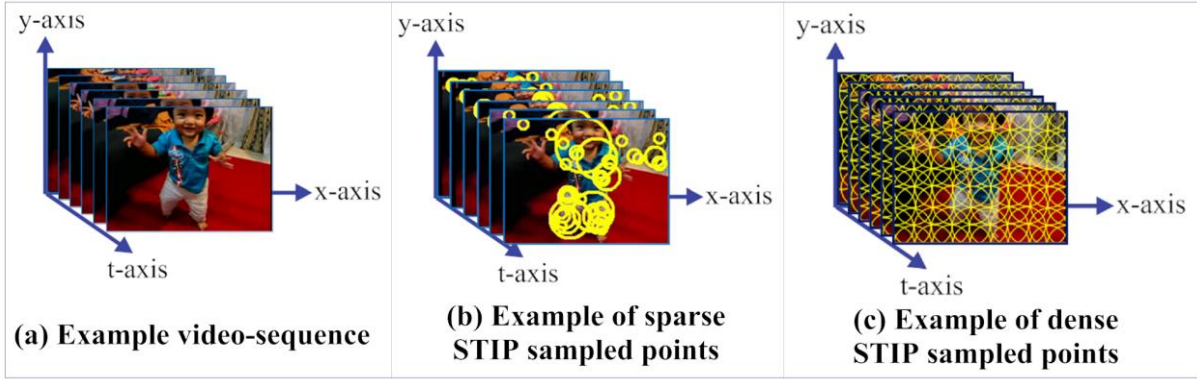

Fig. 2: Example of sampled spatio-temporal interest points (STIP) using sparse STIP detector, Harris3D [1] and dense sampling approach [6]. Circles correspond to different locations at which the STIP were sampled. Varying circles sizes indicate the different scales at which they were sampled.

Fig. 2 depicts the output examples of sparse and dense sampling on a video sequence. Wang et al. [6] in their survey on the performance of local STIP detectors and descriptors, noted that performance of local features are dataset dependent, with most realistic datasets such as the UCF-Sports dataset responds better in terms of classification rate to dense sampling. This is because dense sampling contains background information that can improve classification performance. Human actions can sometime be better defined using its environmental context [17]. Principle of dense sampling involves sampling of video blocks at regular positions and scales in space and time. In their survey, Wang et al. [6] presented that dense sampling is done in a manner in which the minimum size of a 3D patch (i.e. a video block) is 18 pixels in widh and height, and have a span of 10 frames.

We base our choice of detector/ descriptor pairs on the survey done by Wang et al. [6]. The detector/ descriptor pairs that generate most optimal performances for the different datasets are chosen. This includes the usage of Harris3D STIP detector (i.e. sparse type of STIP detector) with Histogram of Oriented Gradient concatenated with Histogram of flow (HOGHOF) [18] for the KTH dataset, and dense sampling paired with histogram of 3D gradient orientations (HOG3D) type of descriptor for the UCF-Sports dataset. The implementation of these methods are available on the authors' websites[1,2]. Default parameters are utilized in our experiments.

In brief, based on the choice detector/descriptors for the different datasets mentioned earlier, having $n$ number of STIP detected, each interest point is mapped to its corresponding local descriptor.

$$P = \{p_1, p_1, ..., p_n\}; D = \{d_1, d_2, ..., d_n\}$$
$$DESC : P \mapsto D$$
(2.1)

Referring to equation (2.1), *DESC* refers to the choice of descriptor for the different datasets as mentioned earlier.

We next perform k-means clustering over the whole set of local descriptors *D*, extracted from the training videos. Clustering is performed to partition the extracted descriptors into *K* number of clusters. Subsequently, we define a set (i.e. a vocabulary) of video-words using the centers of the learned clusters, $\Lambda = \lambda_1, \lambda_2, ..., \lambda_K$. A label, $l$, for a video-word in this context refers to the index value associated with the video word (i.e. for $\lambda_3$, $l$ is then equal to 3), whereby $l \leq K$. Note that $K$ corresponds to the size of the vocabulary created, $|\Lambda|$.

---

[1] http://www.di.ens.fr/~laptev/download.html
[2] http://lear.inrialpes.fr/people/klaeser/software\_3d\_video\_descriptor

## 2.1 Bag-of-Video-Words

At this point we have a set of video-words $\Lambda$, as well as the extracted set of STIP, $P$, along with its corresponding set of local descriptors, $D$. Each of the interest point is first labeled with the label $l$, of its nearest video-word. This is done by computing the Euclidean distance between the descriptor (associated with each interest point) $D$, and the stored video-words $\Lambda$. $K$ as mentioned in the previous section, corresponds to the size of the vocabulary, $|\Lambda|$.

$$Label : P \mapsto P_{labeled}$$
$$p_i \mapsto l_i \;;\; l_i = \underset{j}{\operatorname{argmin}}(dist(d_i, \lambda_j));$$
$$j \in \{1,...,K\}; i \in \{1,...,n\} \tag{2.2}$$

Occurrence of video-words labels in a video is stored in a histogram. This histogram is the bag-of-video-words characterization of a human action contained in a video sequence. This characterization type along with the action labels for the different videos can be used to train and test a SVM based classifier. However, in order to deal with the vocabulary size (as mentioned in Section 1.0), we will next present a video-words selection approach.

## 2.2 Video-Words Selection Using Mutual Information

Often in bag-of-video-words characterization, a large number of video-words are extracted to generalize the dataset containing different actions. For example in the works of Wang et al. , [6], they utilized a vocabulary of the size, $K=4000$ to obtain experimental results.

As our approach utilizes co-occurrence between video-words labels, it is more desirable to have a compact sized vocabulary. A compact sized vocabulary would ease the computation of co-occurrences values between pairs of video-words labels. Note that the calculation of co-occurrences are done for each possible pairs of video-words labels. Mathematically, having $K$ number of video-words labels, the co-occurrence calculation would involve $K^2$ *possible* pairs of video-words labels. If we refer to the example of the works of Wang et al. [6], this means that we have to deal with $K^2=16,000,000$ (i.e. 16 million) pairs of video-words labels. This is computationally very expensive and justifies the need to utilize a method that can reduce the vocabulary size, and at the same time preserves the original classification rate achieved.

Therefore, as part of our proposed approach we have utilized mutual information based video-words selection [19] to reduce the size of the vocabulary, $\Lambda$. This choice is inline with the survey done by Yang et al. [20], that identified mutual information to be among the top performers when used as an evaluation criterion for video-words selection.

Video-words selection is done based on mutual information between the video words, $\Lambda$ and the different action classes, $Y$. The mutual information existing between $\Lambda$ and $Y$ is modeled using the Kullback-Leibler divergence, which is utilized as a distance measure computed between $\Lambda$ and $Y$ across a set of videos containing human actions.

We can treat $\Lambda$ and $Y$ as two discrete random variables. The mutual information between them can be defined as

$$I(\Lambda, Y) = \sum_{\lambda \in \Lambda, y \in Y} p(\lambda, y) \log \frac{p(\lambda, y)}{p(\lambda) p(y)} \tag{2.3}$$

where $p(\lambda, y)$ is the joint distribution of $\Lambda$ and $Y$, $p(\lambda)$ and $p(y)$ are probability distributions of $\Lambda$ and $Y$ respectively. Here the value of the mutual information, $I$ is a dependent variable that depends on the values of both of the independent variables, $\Lambda$ and $Y$.

In our implementation, we utilized the binary logarithm function to realize the computation of mutual information. Using mutual information, we obtain the knowledge of the amount of information from variable $\Lambda$ that is contained in variable $Y$.

During the video-words selection process, a greedy algorithm based on a bottom-up pair wise merging procedure is performed. The algorithm starts by treating each element of $\Lambda$ as a singleton cluster. At each iteration, the elements to be merged are selected based on the pair that causes minimal loss of mutual information.

Assuming $\lambda_1$ and $\lambda_2$ are two candidate elements to be merged, the cost of the merge is defined by the loss of information caused due to this merge, which can be expressed as

$$LI(\lambda_1, \lambda_2) = I(\Lambda_{bef}, Y) - I(\Lambda_{aft}, Y) \qquad (2.4)$$

where $I(\Lambda_{bef}, Y)$ and $I(\Lambda_{aft}, Y)$ correspond to the mutual information before and after the merging in a specific iteration. In our implementation, the merging procedure is achieved by getting the average of each component of the two candidate video-words, i.e.

$$\hat{\lambda}_{ij} = (\lambda_{1j} + \lambda_{2j})/2; i = 1...K^*; j = 1...\ell; \qquad (2.5)$$

where $K^*$ denotes the new compressed vocabulary size and $\ell$ denotes the length of a descriptor corresponding to a video-word.

For extended clarification and the complete mathematics formulation of this technique, we refer the reader to the original work [19]. However, there is still an open issue concerning the size of the reduced vocabulary, $K^*$ (i.e. size of $\hat{\Lambda}$). This will be presented in the following section.

## 2.3 Trade-off Factor for Video-Words Selection

The ultimate goal of mutual information based video-words selection is to find an optimal mapping of video-words from $\Lambda$ to $\hat{\Lambda}$. This is so that the mutual information between $\hat{\Lambda}$ and $Y$ be as high as possible, while the size of $\hat{\Lambda}$ is as compact as possible. Compact sized vocabulary is preferable since it improves the overall computational cost imposed by the bag-of-video-words approach on the final human action classification system.

Naturally, there is a trade off between discriminativeness and compactness in this method of video-words selection. Higher value of mutual information retained at the end of the selection process will lead to $\hat{\Lambda}$ that is more discriminative when used to characterize human actions. On the other hand, although having a more compact (i.e. smaller) size for $\hat{\Lambda}$ is better to reduce the overall computational cost, we risk losing the discriminative power in the retained $\hat{\Lambda}$, that is lost due to the selection process.

To manage this trade-off issue, we define a function that is used to compute a trade-off factor. We denote this factor as $M(\hat{\Lambda})$. The value of this trade-off factor represents a weighted relationship between the reduction of the vocabulary size and the classification rate. Higher reduction in vocabulary size, will produce a higher weighting on the achieved classification rate. Through this measure, we are able to decide on the optimal trade-off between the size of the reduced vocabulary, $|\hat{\Lambda}|$ and the classification rate. A higher value of $M(\hat{\Lambda})$ indicates a better trade-off is achieved, and vice-versa.

$M(\hat{\Lambda})$ is defined using the following equation

$$M(\hat{\Lambda}) = \left(1 - \left(\frac{|\hat{\Lambda}_{reduced}|}{|\hat{\Lambda}_{orig}|}\right)^2\right) \times (\text{ClassificationRate})$$

(2.6)

Referring to equation (2.6), $|\hat{\Lambda}_{reduced}|$ is the size of the reduced vocabulary of video-words, and, $|\hat{\Lambda}_{orig}|$ is the original size of the vocabulary determined in the previous experimental section. The effects of varying the vocabulary sizes will be presented in the experimental section of this paper. The reduced set of video-words $\hat{\Lambda}$ will be the basis from which we will construct our spatio-temporal co-occurrence based characterizations. This will be presented in the following section.

### 3.0 SPATIO-TEMPORAL CO-OCCURRENCE BASED CHARACTERIZATIONS

Fig. 3 depicts the different processes involved in characterizations extraction. These characterizations include spatio-temporal correlations, Haralick texture vector and a PCA co-occurrence vector. These characterizations will be jointly used with the bag-of-video-words approach to characterize human actions.

As specified earlier, the co-occurrence technique will be used along with the reduced vocabulary, $\hat{\Lambda}$. Therefore, the set of vocabulary (i.e. video-words) referred in this section, unless specified, refers to the reduced vocabulary which is $\hat{\Lambda}$. We start this section by presenting the processes involved in constructing spatio-temporal correlograms [10]. These correlograms will be the starting point, in which, different spatio-temporal co-occurrence based characterizations will be extracted.

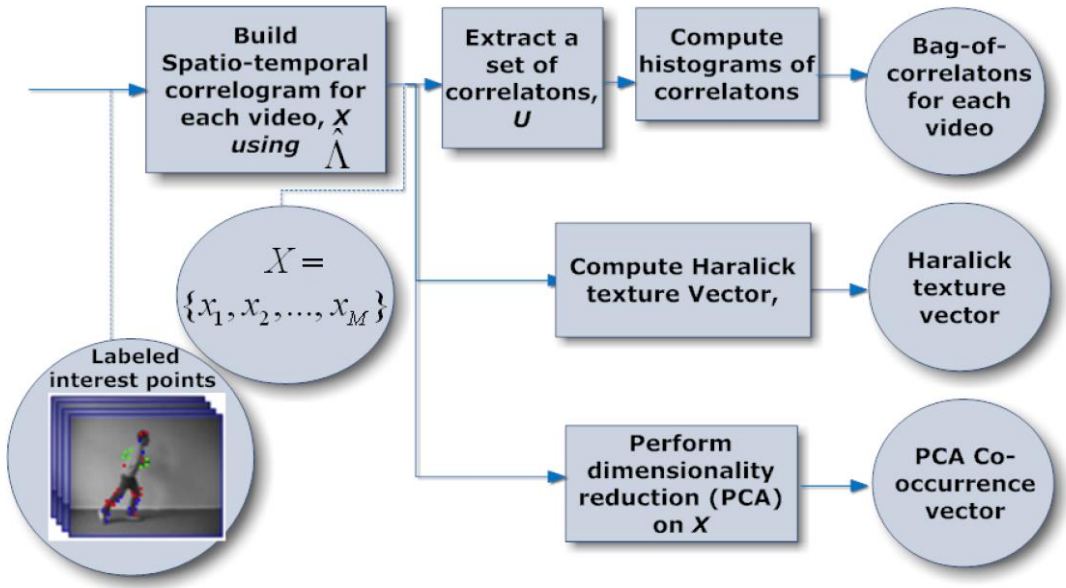

Fig. 3: Action characterization using spatio-temporal co-occurrences

### 3.1 Construction of Spatio-Temporal Correlograms

Following the formation of a reduced set of video-words $\hat{\Lambda}$, similar to the bag-of-video-words approach, across a particular set of videos, in which, the extracted set of STIP, $P$, along with its corresponding set of local descriptors, $D$, each of the interest point will be first labeled with the label of its nearest video-word, $\hat{\Lambda}$.

For each video from the set of videos containing different action classes, we have a distribution of labeled STIP. At this point, we wish to compute the co-occurrences between the video-words labels for each of the videos.

A local histogram $H(\Pi, p)$ is defined as a vector function where each of the vector's component, captures the amount of interest points having a particular label, $l$, that is located within a spatio-temporal kernel $\Pi$, centered on $p$.

For each interest point location, $p$, a set of $J$ kernels centered on $p$ with different sizes is considered. This idea is taken from [10] and uses kernel type (rectangular volume) that extends between *2* to *40* pixels along the spatial dimension and *2* to *60* frames in the temporal domain. The $r^{th}$ kernel of this set is denoted as $\Pi_r$.

The average local histogram is defined as

$$\hat{H}(\Pi_r, l) = \sum_{p \in P_l} \frac{H(\Pi_r, p)}{|P_l|}; 1 \leq r \leq J; 1 \leq l \leq K^* \qquad (3.1)$$

where $P_l$ indicates the set of interest points with label $l$, and $|P_l|$ refers to its cardinality. An average local histogram is computed for each kernel size and for each video-word label as presented in equation (3.1). A *correlogram*, $x$, for a particular video, $Vidx$, is built by concatenating in an array such average local histograms for all combinations of labels and kernels.

$$x = \begin{bmatrix} \hat{H}(\Pi_1, 1) & \cdots & \hat{H}(\Pi_1, K^*) \\ \vdots & \ddots & \vdots \\ \hat{H}(\Pi_J, 1) & \cdots & \hat{H}(\Pi_J, K^*) \end{bmatrix} \qquad (3.2)$$

The schematic of constructing a spatio-temporal correlogram is presented in Fig. 4.

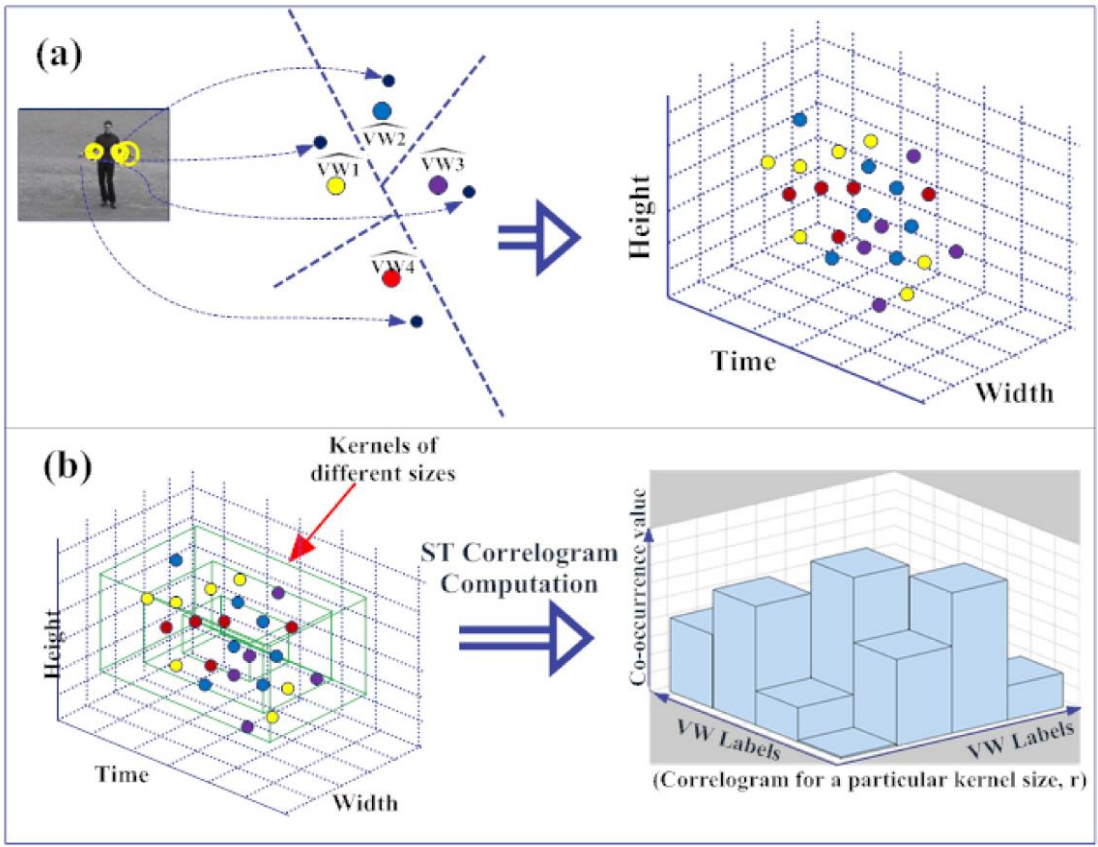

Fig. 4: Schematic of spatio-temporal correlogram construction. (a) For each video, interest points are first labeled with their nearest reduced video-words producing a distribution of labeled interest-points. (b) Spatio-temporal relationship between labeled STIP are stored in a spatio-temporal correlogram constructed using kernels of varying sizes

At the end of this process, we will obtain a set of spatio-temporal correlograms for a set of videos containing different actions, $X = \{x_1, x_2, ...x_M\}$, with $M$ being the number of videos in a particular dataset. A correlogram contains the co-occurrences between video-words labels for a video containing human action. A spatio-temporal correlogram is dimensionally large, of the size $K^* \times K^* \times J$. Therefore, the next step involves extracting useful spatio-temporal co-occurrence information from the correlograms in order to utilize it for characterizing human actions.

### 3.2 Extraction of Spatio-Temporal Correlations

Referring to the previous definition of a spatio-temporal correlogram in equation (3.2), we first observe a particular label, $a$ (which is a column of the array defined by equation (3.2)). We then define $x_a$ as a vector of the average local histograms for label $a$, $\hat{H}(\Pi, a)$, where each component characterizes a particular size of the kernel used, $\Pi_i$, $i = 1,...,J$. And so, $x_a$ is comprised of average local histograms for label $a$, for all kernel sizes. Note that $K^*$ is the vocabulary size (i.e. the total number of labels).

$$x_a = \begin{bmatrix} \hat{H}(\Pi_1, a) \\ \vdots \\ \hat{H}(\Pi_J, a) \end{bmatrix}; a \leq K^* \tag{3.3}$$

Defining $\hat{H}_b(\Pi, a)$ as a particular value of histogram $\hat{H}(\Pi, a)$, at bin $b$ that corresponds to a specific video-word, we can further expand the definition for $x_a$.

$$x_a = \begin{bmatrix} \hat{H}_1(\Pi_1, a) & \cdots & \hat{H}_b(\Pi_1, a) & \cdots & \hat{H}_K(\Pi_1, a) \\ \vdots & & \ddots & & \vdots \\ \hat{H}_1(\Pi_J, a) & \cdots & \hat{H}_b(\Pi_J, a) & \cdots & \hat{H}_K(\Pi_J, a) \end{bmatrix}; a \leq K^*; b \leq K^* \tag{3.4}$$

Therefore a correlogram element, $v(a, b, Vidx)$ for video $Vidx$, can be defined as a vector containing the values of $[\hat{H}_b(\Pi_1, a),..., \hat{H}_b(\Pi_J, a)]$. A correlogram element represents the spatio-temporal co-occurrence between video-words labels $a$ and $b$. The spatio-temporal correlogram elements are then collected across a set of videos, to form a set of correlogram elements, $V = \{v_1, v_2,..., v_G\}$. The number of correlogram elements, $G$, is of the size $K^* \times K^* \times M$, with $M$ is the number of videos existing in a particular dataset.

K-means clustering is then performed on the whole set of correlogram elements, $V$, to partition it into $Q$ clusters, $C\_corr = \{c\_corr_1, c\_corr_2...c\_corr_Q\}$. A set of spatio-temporal correlations is then defined as the centers of these clusters, denoted by $U = \{u_1, u_2,..., u_Q\}$. We utilize the correlations to construct a bag-of-correlations characterization for each video sequence.

## 3.3 Extraction of Spatio-Temporal Haralick Texture Vector

In this section, we will present another type of characterization for human actions based on Haralick texture measures [14]. We consider the usage of Haralick texture measures based on the distribution of labeled video-words. Specifically, we wish to extract spatio-temporal texture information of different action classes, that is embedded within the distribution of the labeled video-words from each video containing a particular action. Haralick introduced 13 different measures to extract texture information from 2D images. Details concerning the formula for the different measures can be referred to the original paper [14].

Building blocks of a spatio-temporal correlogram are local histograms of varying kernel sizes. Again, referring to the definition for a spatio-temporal correlogram in equation (3.2) and by observing a particular kernel index, $j$, we can define $x_{\Pi j}$, which is essentially an array of local histograms for all labels for a specific kernel size, $j$. More specifically, if we refer to row components of the previous definition from equation (3.2), we can define $x_{\Pi j}$ as

$$x_{\Pi j} = [\hat{H}(\Pi_j, 1), ..., \hat{H}(\Pi_j, K^*)]; j \leq J \qquad (3.5)$$

Typically, Haralick texture measures are extracted from 2D correlograms. Referring to the previous equation, in our work, a 2D correlogram for a particular kernel size, $j$ (denoted as $x_{\Pi j}$), is an array of local histograms stacked together, to form a 2D correlogram of the size $K^* \times K^*$. We then proceed by extracting Haralick texture measures from each of the 2D correlogram, $x_{\Pi j}$.

We define a mapping from a correlogram, $x$ to its corresponding Haralick texture vector denoted as $\beta$. The mapping involves the computation of the Haralick texture measures to form $\beta$ for each $x$

$$\begin{aligned} Haralick : x \mapsto \beta \\ x_{\Pi j} \mapsto \beta_m \\ j \in \{1, ..., J\}; m \in \{1, ..., 13\} \end{aligned} \qquad (3.6)$$

The Haralick texture vector, $\beta$, which has $13 \times J$ elements, is formed by concatenating each $\beta_m$ extracted from each $x_{\Pi j}$ into a single vector.

This, along with other characterization types discussed can be used in combination or separately to represent and characterize each video sequence. We will next present another type of novel co-occurrence based characterization.

## 3.4 Extraction of PCA Co-occurrence Vector

In this section, we explain how we propose to convert the spatio-temporal correlogram defined in equation (3.2) into its corresponding PCA co-occurrence vector. The goal of this approach is to avoid vector quantization that risks loss of information between the co-occurrences of video-words labels.

Currently, we have a set of spatio-temporal correlograms extracted from a set of videos containing different actions, $X = \{x_1, x_2, ... x_M\}$, with $M$ being the number of videos in a particular dataset.

A single spatio-temporal correlogram, $x_i$ defined by equation (3.2) is of the size $K^* \times K^* \times J$, with $K^*$ being the number of video-words labels and $J$ being the number of different kernel sizes. By treating each $x_i$ as a 1-

dimensional vector defined as $vec\_x_i$, we will have a collection of 1-d vectors, defined as $VEC\_X$, that is extracted from a particular dataset with $M$ number of videos.

Given a collection of these 1-D vectors, $VEC\_X = \{vec\_x_1, vec\_x_2, ..., vec\_x_M\}$, we proceed by performing PCA to reduce the dimensionality of each $vec\_x_i$. In our work, this is done by fixing the number of principal components, $S$, with $S << (K^* \times K^* \times J)$, and by projecting each elements of $VEC\_X$ onto a new orthogonal basis, creating a new set of dimensionally reduced vectors, $Z = z_i, ..., z_M$. This allows the reduction of dimensions and preservation of important information. The projected vectors, are smaller in dimension and contain essential co-occurrence information that can be used to characterize videos containing human actions.

These PCA Co-occurrence vectors that characterize different human actions in different video sequences, will be used to train and test an SVM based classifier. SVM enables us to use the 'kernel-trick' [21], allowing us to map the originating features (i.e. different characterization approaches such as bag-of-video-words, Haralick texture vector and PCA Co-occurrence vector), into a higher-dimensional feature space. To fully exploit the discriminative power contained within each characterization type, we propose the usage of multiple-channel kernel based SVM that utilizes different kernels for different types of characterizations.

## 4.0 COMBINATION OF CHARACTERIZATIONS: MULTIPLE CHANNEL KERNEL BASED SVM

We now have multiple characterizations for human actions as presented in Table 1. In SVM [21], kernel trick is used to map the original set of observations into an inner product space, without ever having to compute the mapping explicitly. It is almost intuitive to consider the combination of different kernels as opposed to make a choice of one single most optimal kernel for different types of characterizations.

Table 1: Different characterizations of human actions

| Bag-of-Video-Words (BoVW) | Bag-of-Correlations (BOC) | Haralick Texture Measures (Hara) | PCA Co-occurrence Vector (PCA-Cooc) |
|---|---|---|---|
| Histogram of occurrences of video-words labels | Histogram of vector quantized spatio-temporal correlogram elements | Thirteen texture measures [14](e.g. Energy, Entropy, Inertia and Correlation) | Dimensionally reduced spatio-temporal correlogram vector |

We utilized a multiple-channel kernel based SVM [22] to combine different types of characterizations. Each channel, $c$, in our case refers to the different types of characterizations such as bag-of-video-words, bag-of-correlations, Haralick texture vector and PCA Co-occurrence vector. Here, different distance measures are utilized to combine the different channels (i.e. characterizations), and the kernel, $K_{comb}$, containing the combined distance measures between different channels are used to train and test an SVM classifier.

The combination kernel, $K_{comb}$ can be mathematically explained as

$$K_{comb}(x, y) = \exp(-\sum_c \frac{1}{\Omega_c} D(x^c, y^c)) \tag{4.1}$$

where $D(x^c, y^c)$ is the distance computed using channel $c$ and $\Omega_c$ is the normalization factor computed as an average channel distance [22], and $x, y$ is a pair of videos each containing a particular human action. In our implementation, we selected $\chi^2$ distance for both the histogram based characterizations (i.e. for bag-of-video-words histogram vector, and bag-of-correlations histogram vector). For the other two (i.e. Haralick texture vector and PCA Co-occurrence vector) characterizations, we utilized L2 (euclidean) distance.

We have now presented all characterization types that are proposed within this paper. We have also presented the technique on which we have utilized to combine and classify using the different types of characterizations. Therefore, the next section will focus on the experimental results to evaluate our proposed approach to characterize human actions.

## 5.0     EXPERIMENTAL RESULTS

The experimental results are obtained through the usage of two distinct and highly challenging datasets, the KTH and UCF-Sports dataset. Both of these datasets were also chosen to provide similar benchmarks for our proposed approach with other methods that perform human action classification such as [18][6][23].

For both datasets, we first extract a set of local features by utilizing the specified pairs of STIP sampling and description approach presented in Section 3. We randomly select 100,000 local features from the set of local features that is extracted from training videos of each dataset [6]. This will be the basis in creating the vocabulary of video-words. However, instead of fixing the number of the initial video-words, $\Lambda$ (size of initial vocabulary) to 4000 as in [6], we fixed the number of initial video-words to 1000 which has been shown in our experiments to produce consistently good results for various action datasets. Next, we will present the characteristics of the two datasets along with the initial classification results obtained using the bag-of-video-words characterization approach.

KTH dataset [16] contains 6 types of actions. There are 599 low-resolution (160 x 120) video files from a combination of 25 subjects, 6 actions and 4 scenarios (scenarios consist of variations in lighting, scale as well as videos taken indoor and outdoor). In our experiments, we followed the original experimental setup of the authors [16]. The dataset is divided into a test set (9 subjects), while the rest (16 subjects) are used as training set. Initial size of the vocabulary, $\Lambda$ used to create the bag-of-video-words histogram vector is 1000, in which we obtained **91.67%** classification rate.

The UCF-Sports dataset is a collection of 150 broadcast sports videos and contains 10 different actions. It is a highly challenging dataset with large variations in terms of scenes and viewpoints. We adopted the experimental setup utilized by [24] that utilizes a 5-fold-cross-validation setup. To increase the number of training data, we followed the steps of [6] by flipping the videos from left to right. The initial size of the vocabulary, $\Lambda$ used to create the bag-of-video-words histogram vector is 1000, in which we obtained **86.67%** classification rate.

We will first present an experiment on the effects of varying the size of the reduced vocabulary, $\hat{\Lambda}$ that is obtained using the mutual information based clustering discussed in Section 3. Next, we present an experiment that is conducted to discover the optimal number of spatio-temporal correlations based on its effect on the classification performance. Similarly, different sizes of PCA Co-occurrence vector are evaluated in order to determine its optimal size. Finally, we present the performance of our proposed approach on the overall classification of human actions, by studying the effects of combining the proposed characterizations using multi-channel kernel based SVM.

### 5.1     Results for Video-Words Selection

In this section we present the results of classification performance by varying the size of the reduced vocabulary $\hat{\Lambda}$, using an initial vocabulary $\Lambda$ created of the size 1000. The vocabulary size is reduced utilizing a mutual information approach (refer Section 2.2). This reduced vocabulary $\hat{\Lambda}$, is then used to create a bag-of-video-words histogram vector. Next, an SVM based classifier is trained and tested. The results for KTH and UCF-Sports dataset is presented in Table 2.

Table 2: Effects of the reduced vocabulary sizes on the classification rate for KTH and UCF-Sports datasets, and their associated $M(\hat{\Lambda})$ factors

| Size of $\hat{\Lambda}$ | Classification rate % KTH | $M(\hat{\Lambda})_{KTH}$ | Classification rate % UCF-Sports | $M(\hat{\Lambda})_{UCF}$ |
|---|---|---|---|---|
| 100 | 85.19 | 84.34 | 69.66 | 68.96 |
| 200 | **88.89** | **85.33** | 74.49 | 71.51 |
| 300 | 83.80 | 76.26 | **83.04** | **75.57** |
| 400 | 88.65 | 71.94 | 80.49 | 51.51 |
| 500 | 88.89 | 66.67 | 79.20 | 59.40 |
| 600 | 89.35 | 57.18 | 82.46 | 52.77 |
| 700 | **91.67** | 46.75 | **87.01** | 44.38 |
| 800 | 89.35 | 32.17 | 84.26 | 30.34 |
| 900 | 89.81 | 17.06 | 84.36 | 16.03 |
| 1000 | 91.67 | 0 | 86.67 | 0 |

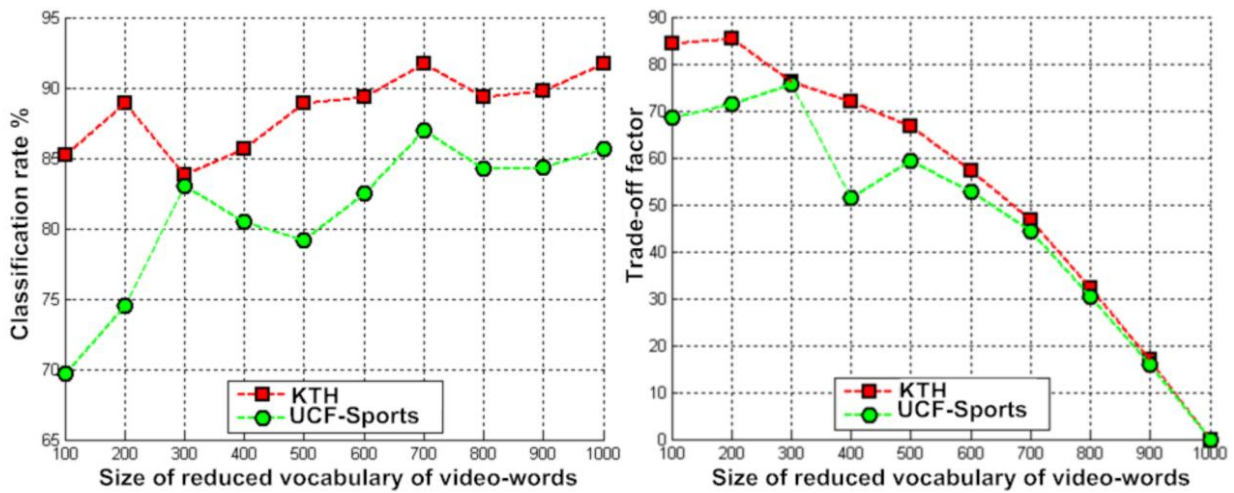

Fig. 5: (Left) Classification rate vs. size of reduced vocabulary of video-words, (Right) Trade-off factor vs. size of reduced vocabulary of video-words

Referring to Fig. 5, for the KTH dataset, the classification rate reaches several local maxima. However, the most prominent peaks are those when the number of video-words are reduced to 200 and 700. The highest value of $M(\hat{\Lambda}) = 85.33$ is when $|\hat{\Lambda}| = 200$. This indicates that we are able to select 200 (preserving **88.89%** classification rate) as the size of the reduced video-words, $\hat{\Lambda}$, for the KTH dataset, obtaining an optimal trade-off between the reduction in vocabulary size and the classification rate. This allows us to reach our objective, that is, to be able to utilize low number of video-words and promote improvement of classification rate using spatio-temporal co-occurrence technique since the size of the vocabulary of video-words is greatly reduced.

Similarly, for the UCF-Sports dataset, there are several local maximas, with the most prominent peaks to be $\hat{\Lambda} = 300$ and $\hat{\Lambda} = 700$. Similar to the reasons stated previously, we utilize the value of $M(\hat{\Lambda}) = 75.57$ to determine the optimal trade-off between the reduction in vocabulary size and its effect on the classification rate. Therefore, we are able to select 300 as the size of $\hat{\Lambda}$, for the UCF-Sports dataset, preserving **83.04%** classification rate.

The results presented in this section demonstrate that our overall target of obtaining a reduced size for the vocabulary $\hat{\Lambda}$, while preserving the classification performance is achieved. This is demonstrated by the fact that we are able to maintain more than 80% classification rate even when reducing the size of the vocabulary between 20% to 30% of its initial size.

### 5.2 Optimal Number of ST Correlations

Here, we will first present the experimental results of varying the number of correlations on the classification rate using the optimal sizes of the reduced set of vocabulary $\hat{\Lambda}$ presented in the previous section. Bag-of-correlations are used to train and test an SVM based classifier. Results for both the KTH and UCF-Sports dataset are presented in Table 3.

Table 3: Varying sizes of correlations and their effects on the classification rate for KTH and UCF-Sports dataset

| Size of correlations | 100 | 200 | 300 | 400 | 500 |
|---|---|---|---|---|---|
| Classification rate, %, KTH | 61.57 | 66.20 | 62.96 | **67.59** | 65.74 |
| Classification rate, %, UCF | 69.38 | 69.83 | **71.88** | 68.66 | 69.53 |

We can observe from Table 3 that classification rates are at their maximum when size of correlations are 400 (for KTH) and 300 (for UCF-Sports). It is evident that the usage of correlations to a certain extent is able to classify different types of human actions, achieving near **70%** classification rate for both datasets. However, if we perform a direct comparison with bag-of-video-words characterization that obtained more than **80%** classification rate for both datasets, we can deduce that usage of correlations produces lower classification performance. This is inline with our hypothesis. Eventhough a spatio-temporal correlogram is able to record co-occurrence information, usage of vector quantization on correlogram elements contributes to the lost of important information, as the information concerning video-words labels is lost during vector quantization. This justifies the importance of finding alternatives to correlations.

We will next present the effect of varying the size of PCA Co-occurrence vector on the classification performance.

**5.3    Optimal size for the spatio-temporal PCA co-occurrence vector**

We will present a study on the relationship between varying sizes $S$, of the PCA Co-occurrence vector, and the classification rate achieved. This experiment is done to identify the optimal size of the PCA Co-occurrence vector on the basis of the size that generates the highest classification rate. Varying sizes of the PCA Co-occurrence vector are used to train and test a SVM based classifier.

It is important to note that the number of training video sequences $N_{trainPCA}$, for the KTH dataset is 383, and an average of 238 video sequences for the UCF-Sports dataset. We tested different values for $S$ with $N_{trainPCA} - 1$ [25] as its maximum value. The results for both the KTH and UCF-Sports dataset are as denoted in Table 4. Note that N/C in the table stands for 'Not Computed'.

Table 4: Varying sizes, $S$ of PCA Co-occurrence vector, and their effects on the classification rate for KTH and UCF-Sports dataset

| Size of $S$ | 10 | 50 | 100 | 150 | 200 | 237 | 250 | 300 | 350 | 382 |
|---|---|---|---|---|---|---|---|---|---|---|
| KTH | 75.46 | 85.65 | 86.57 | 86.57 | **87.96** | N/C | 87.96 | 87.96 | 87.04 | 87.96 |
| UCF-Sports | 53.75 | 72.17 | **81.12** | 81.12 | 81.12 | 81.12 | N/C | N/C | N/C | N/C |

We can observe from Table 4 that classification rates are at their maximum when size of PCA Co-occurrence vectors $S$, are 200 (for KTH) and 100 (for UCF-Sports). Increasing $S$ beyond these values does not introduce any classification rate improvement. Using a value of $S = 50$ for both datasets, we achieve more than **70%** classification rate. This indicates that usage of PCA to produce the PCA-Cooc characterization vector managed to retain important information from the correlogram even when its dimension is greatly reduced. Using the optimal sizes for the correlations and PCA Co-occurrence vector, we will next study the effect of combining the different characterizations presented.

## 5.4 Combination of Different Characterizations

In this section, different characterizations for human actions which are based on bag-of-video-words (BoVW), bag-of-correlations (BOC), Haralick texture vector (Hara) and PCA Co-occurrence vector (PCA-Cooc) and their possible combinations will be tested. In essence, this experiment is to determine which possible combination gives the best classification result and whether the proposed approach is successful in exploiting the spatio-temporal co-occurrence information embedded within the spatio-temporal correlograms to augment the performance of the standard bag-of-video-words approach.

We will next present the results of the performance of individual characterizations of human actions, along with their combinations presented in this paper in Table 5. Note that $BoVW_{orig}$ refers to bag-of-video-words characterization generated using the optimal size of vocabulary obtained using K-means clustering. Meanwhile, $BoVW_{MI}$ refers to the bag-of-video-words characterization generated using mutual information based video-words selection.

Table 5: Comparison table - Individual characterizations (left), their combinations (right) and their effects on the classification rate (%) for KTH and UCF-Sports dataset

| Individual Characterization Type | KTH | UCF-Sports |
|---|---|---|
| $BoVW_{orig}$ | 91.67 | 86.67 |
| $BoVW_{MI}$ | 88.89 | 83.04 |
| BOC | 67.59 | 71.88 |
| PCA-Cooc | 87.96 | 81.12 |
| Hara | 62.04 | 65.16 |

| Combination of characterizations | KTH | UCF-Sports |
|---|---|---|
| $BoVW_{MI}$ BOC | 89.35 | 81.63 |
| $BoVW_{MI}$ PCA-Cooc | 92.13 (Best result for KTH) | 89.09 |
| $BoVW_{MI}$ Hara | 88.89 | 88.69 |
| $BoVW_{MI}$ BOC PCA-Cooc | 92.13 | 87.08 |
| $BoVW_{MI}$ BOC Hara | 87.04 | 82.41 |
| $BoVW_{MI}$ PCA-Cooc Hara | 92.13 (Best result for KTH) | 91.30 (Best result for UCF) |
| $BoVW_{MI}$ BOC PCA-Cooc Hara | 88.43 | 87.79 |

For the **KTH** dataset, based on the results presented in the comparison table (i.e. Table 5), we can deduce that among the three different spatio-temporal co-occurrence based characterizations (BOC, Hara and PCA-Cooc), our proposed spatio-temporal PCA co-occurrence vector (i.e. 'PCA-Cooc') achieves the highest individual classification rate at **87.96%**. In fact, when compared side by side with the $BOVW_{MI}$ performance, the classification performance is rather close (less than **1%** difference for KTH). Results in the comparison table also demonstrates that, PCA-Cooc is able to improve the classification rate beyond the usage of standard $BoVW_{orig}$ approach. The combination of $BoVW_{MI}$ and PCA-Cooc, is able to slightly improve the classification rate to **92.13%**, in comparison to using $BoVW_{orig}$ alone (**91.67%**). Other combinations do not exceed the classification improvement provided by the combination of $BoVW_{MI}$ and PCA-Cooc.

The classification improvement introduced in the KTH dataset is minimal, and therefore it is evident that we need to analyze the performance of our proposed approach on the **UCF-Sports** dataset that is more realistic in nature. Fig. 6 contains some classification result obtained using a SVM classifier that is trained solely on the KTH dataset, and tested on (our) recorded videos that does not belong to the KTH dataset.

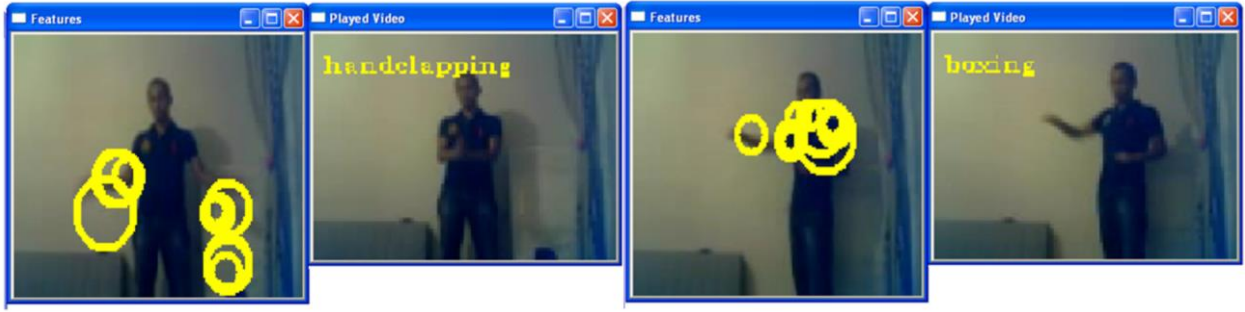

Fig. 6: Example of classification results obtained using SVM classifier trained on KTH training data, utilizing combined characterizations that produces highest classification rate for KTH (refer Table 5). Successful classification on low resolution videos ($160 \times 120$), in low-light conditions.

For the UCF-Sports dataset, results indicate similar trends. 'PCA-Cooc' is observed to achieve the highest classification rate when compared with other co-occurrence based characterizations such as Hara and BOC. Classification rate improvement is obtained when the $BoVW_{MI}$ vector is combined with the 'PCA-Cooc' vector obtaining **89.09%** classification rate. However, an even higher classification rate is obtained by the combination of $BoVW_{MI}$, PCA-Cooc and Hara, in which we obtained **91.3%** classification rate. This indicates that both PCA-Cooc and Hara were able to extract relevant spatio-temporal co-occurrence information from the spatio-temporal correlograms and increases the overall classification rate for the UCF-Sports dataset.

PCA-Cooc as demonstrated for both KTH and UCF-Sports is the most discriminative out of all the proposed spatio-temporal co-occurrence based characterization for human actions reaching **87.96%** for KTH and **81.12%** for UCF-Sports. Its performance is **20%** higher than other types of co-occurrence based characterizations when used individually. Even though Hara does not perform well when used individually, it introduces improvement when used in combination with PCA-Cooc. This is demonstrated through the highest classification rate achieved in our experiment for the UCF-Sports dataset, when using the combination of $BoVW_{MI}$, PCA-Cooc and Hara (**91.3%**). This implies that Hara was successful to a certain extent in measuring the co-occurrence information contained in a spatio-temporal correlogram.

Using BOC alone or in combination with other characterizations fail to surpass the classification rate improvement provided by PCA-Cooc and Hara. This supports our initial hypothesis, in which video-words label information is essential in preserving the spatio-temporal co-occurrence information contained in a spatio-temporal correlogram. This is mainly because the BOC does not correctly preserve video-words labels information. This information is lost during the vector quantization process (i.e. K-means clustering) that was performed to generate the set of spatio-temporal correlations.

Table 6: Classification results for the KTH dataset

| Characterization/ Action | $BoVW_{orig}$ (%) | Combined $BoVW_{MI}$, PCA-Cooc and Hara (%) |
|---|---|---|
| **Box** | 100.00 | 100.00 |
| **Clap** | 97.22 | 97.22 |
| **Wave** | 91.67 | 94.44 |
| **Jog** | 83.33 | 94.44 |
| **Run** | 77.77 | 66.67 |
| **Walk** | 100.00 | 100.00 |
| **Average accuracy** | 91.67 | 92.13 |

Table 6 contains the average classification rate for each action class for the KTH dataset. The combination of characterizations that contributes to highest classification rate is compared class-by-class with the classification rate performance based on $BoVW_{orig}$ characterization. For the KTH dataset, we achieved improvement for the Wave and Jog action classes, with the improvement ranging between **+2.77%** to **+11.11%**. The overall classification improvement for the KTH dataset is rather minimal from **91.67%** to **92.13%**. Therefore, as noted before it is important to evaluate our approach against a more realistic dataset such as the UCF-Sports dataset.

Table 7: Classification results for the UCF-Sports dataset

| Characterization/ Action | $BoVW_{orig}$ (%) | Combined $BoVW_{MI}$, PCA-Cooc and Hara (%) |
|---|---|---|
| Dive | 100.00 | 100.00 |
| Golf | 77.78 | 94.44 |
| Kick | 94.74 | 100.00 |
| Lift | 100.00 | 100.00 |
| Ride | 66.67 | 66.67 |
| Run | 84.62 | 76.92 |
| Skate | 75.00 | 91.67 |
| Swg-bench | 100.00 | 95.00 |
| Swg-side | 100.00 | 92.31 |
| Walk | 68.18 | 95.45 |
| Average accuracy | 86.67 | 91.3 |

Similarly, Table 7 contains the average classification rate for each action class for the UCF-Sports dataset. Class-by-class comparison in terms of classification performance is done between the combination of characterizations with the highest classification rate with $BoVW_{orig}$ characterization. For the UCF-Sports dataset, improvement in terms of classification performance is experienced in numerous action classes such as **Golf: +16.67%**, **Kick: +5.26%**, **Skate: +16.67%** and **Walk: +27.27%**. There is a slight decrease in terms of performance for 3 different action classes (i.e. Run, Swing-bench and Swing-side), that falls in the range between **5.0%** to **7.7%**. This suggest that in some action classes, the usage of co-occurrence information might amplify the existing noise (due to occlusion, background clutter etc.), and reduce the classification performance. However, this minimal decrease in performance is compensated by the classification improvement obtained.

In general, our proposed approach is able to improve the overall the classification rate across different action classes with significant classification rate improvement varying between **5.0%** to almost **30.0%** when tested against a realist dataset such as UCF-Sports. We will next discuss the performance of our proposed approach against a number of state-of-the-art approaches.

## 6.0   COMPARISON WITH STATE-OF-THE-ART RESULTS

Table 8: Comparison table, state-of-the-art for KTH (left) and UCF-Sports (right) datasets

| KTH | Year | Classification Rate (%) | UCF-Sports | Year | Classification Rate (%) |
|---|---|---|---|---|---|
| Laptev et al. [18] | 2008 | 91.8 | Wang et al. [6] | 2009 | 85.6 |
| Kovashka and Grauman [23] | 2010 | 94.53 | Kovashka and Grauman [23] | 2010 | 87.27 |
| Gilbert et al. [26] | 2011 | 94.5 | O'Hara and Draper [27] | 2012 | 91.3 |
| (Ours) Combined $BoVW_{MI}$, PCA- | 2015 | 92.13 | Ravanbakhsh et al. [28] | 2015 | 97.8 |

| Cooc and Hara | | |
|---|---|---|
| (Ours) Combined BoVW$_{MI}$, PCA-Cooc and Hara | 2015 | 91.3 |

Referring to Table 8, our method is able to challenge the current state-of-the-art performance for both of the datasets. For the KTH dataset, the classification rate achieved is rather similar to the performance achieved by [18]. The difference in terms of classification performance of our approach in comparison to other approaches in Table 8 (besides [18]) is rather minimal, ranging in between 1 and slightly more than 2 percent. This is due to the fact that the KTH dataset is a simple dataset and the only way to properly analyze the performance of our proposed approach is to test it against a more realistic dataset such as the UCF-Sports dataset.

For the UCF-Sports data-set we are able to equalize the classification rate achieved by O'Hara and Draper [27] and achieve a significant performance when compared against other state-of-the-art works. However, it is important to note that their classification performance is partly contributed by the fact that their work does not utilize any feature extraction approach. Instead they depended on optimal bounding boxes provided with the UCF-Sports dataset [5]. It is also important to note that although our method is lesser in terms of performance in comparison to the works of Ravanbakhsh et al. [28], their performance is obtained via an elaborate approach consisting of convolutional neural networks (i.e. 'deep learning') which is known to provide enhanced performance. Our method is able to challenge their performance via a more straight forward feature-centric approach.

## 7.0 CONCLUSION AND FUTURE WORKS

In our work, we proposed the usage of a trade-off factor in a video-words selection method to increase the efficiency in the creation of spatio-temporal correlograms. From these correlograms, we extracted three novel co-occurrence based characterizations (i.e. bag-of-correlations, bag-of-video-words and Haralick texture vector). The main aim is to make full use of the co-occurrence information to characterize human actions. Our result demonstrates that the proposed PCA Co-occurrence vector, and Haralick texture vector, contribute to the best classification rate when used in combination with standard bag-of-video-words characterization. We also validated our hypothesis based on experimental results, that video-words label information is essential in ensuring co-occurrence information is correctly preserved. We believe the different characterizations have distinctive discriminative power, and the key to exploiting them is to ensure successful combination. Future works include experimenting with different feature combination techniques to fully take advantage of the different characterizations that have been proposed.